\documentclass[letterpaper, 10 pt, conference]{ieeeconf}  

\IEEEoverridecommandlockouts                              

\usepackage[utf8]{inputenc}

\usepackage{xspace}
\usepackage{comment}
\usepackage{graphicx}
\usepackage{xcolor}
\definecolor{gg}{RGB}{0, 155, 85}
\definecolor{primarycolor}{RGB}{33,49,77}   
\definecolor{angrycolor}{RGB}{210,73,42}    

\usepackage{todonotes}

\usepackage{amsmath}
\usepackage{amssymb}
\usepackage{amsthm}
\theoremstyle{definition}
\usepackage{tabularx}

\usepackage{tikz}
\usetikzlibrary{automata,positioning,angles,quotes}
\usepackage{tikz-cd}

\makeatletter
\let\NAT@parse\undefined
\makeatother
\usepackage[pdfa,colorlinks,bookmarksopen,bookmarksnumbered,allcolors=gg]{hyperref}
\usepackage{cite}

\PassOptionsToPackage{ruled,vlined,linesnumbered}{algorithm2e} 
\usepackage{algorithm2e}
\RestyleAlgo{ruled}   
\LinesNumbered        
\AtBeginEnvironment{algorithm}{\small}      
\SetKwComment{tcp}{\quad\color{gray!70}// }{} 
\SetKwInput{KwIn}{Input}
\SetKwInput{KwOut}{Output}

\SetAlgoSkip{SkipBeforeAndAfter}

\SetCommentSty{mycommfont}
\SetKw{Continue}{continue}
\usepackage{booktabs}
\usepackage{multirow}
\usepackage{makecell}

\usepackage[font=footnotesize]{caption}
\usepackage[font=footnotesize]{subcaption}
\usepackage[export]{adjustbox}
\usepackage{booktabs, tabularx, makecell, multirow}
\usepackage{mathtools}

\usepackage[
activate   = {true},
protrusion = true,
expansion  = true,
kerning    = true,
spacing    = true,
tracking   = false,
auto       = true,
selected   = true,
factor     = 1000,
stretch    = 15,
shrink     = 15,
]{microtype}

\newcommand{\lsr}{\textsc{lsr}\xspace}

\usepackage{color}

\usepackage{xspace} 

\newcommand{\Vis}{\ensuremath{\mathcal{J}}\xspace}
\usepackage{booktabs}
\usepackage{multirow}
\usepackage{makecell}

\title{\LARGE \bf
Learning Discrete Abstractions for Visual Rearrangement Tasks Using Vision-Guided Graph Coloring 
}
\author{Abhiroop Ajith and Constantinos Chamzas}

\begin{document}

\maketitle
\thispagestyle{empty}
\pagestyle{empty}


\begin{abstract}
Learning discrete abstractions directly from visual data is a central challenge in robotics. While classical task-planning methods assume a predefined symbolic state space, robots operating from raw RGB observations must first infer such a structure before planning becomes possible. In this work, we study abstraction learning in visual rearrangement tasks, where the robot is provided only with image-based execution traces and high-level pick-and-place action primitives.
We propose a constrained graph-construction framework that induces discrete task graphs directly from visual data. Our method enforces structural properties inherent to rearrangement domains, such as bipartiteness and action uniqueness, thereby restricting the search to structurally valid abstractions. To select among feasible candidates, we introduce an attention-guided optimal-transport visual distance that promotes clustering of task-equivalent observations without object-level supervision.
To enable end-to-end planning from novel images, we additionally learn an image to abstract node assignment model leveraging the clusters as labels, allowing start/goal images to be localized on the learned graph at test time. 
We evaluate our method across seven simulation benchmarks and two real robot experiments with a UR10 robot mounted with an Intel RealSense Camera. Our approach produces structurally consistent task graphs and achieves higher planning success compared to representation learning clustering baselines, with $94\%$ and $70\%$ image-to-plan success rate on real data. 
\end{abstract}



\section{Introduction}






A central challenge in artificial intelligence and robotics is enabling agents to reason at an abstract task-relevant level while operating in continuous, high-dimensional sensorimotor spaces. Humans routinely bridge this gap~\cite{mason_toward_2018}: when packing a lunch, one plans in terms of subgoals—open the backpack, place the lunch inside, zip it close—rather than reasoning over individual joint torques or pixels. This separation between high-level symbolic reasoning and low-level motor execution enables efficient long-horizon behavior and is fundamental to hierarchical cognition and action control~\cite{botvinick2014model}.

Indeed, once a suitable discrete abstraction is available, long-horizon decision making reduces to graph search over a finite state space, yielding solutions that are efficient, interpretable, and verifiable. This is the domain of classical task planning, studied extensively in artificial intelligence~\mbox{\cite{blum1997fast,helmert2006fast}}. Crucially, however, these methods presuppose that the symbolic state space is given: discrete state variables and operator schemas must be specified a priori \cite{helmert2006fast}. When a robot perceives the world through raw image observations, this structure is not directly available. The core difficulty is therefore in inferring the task-level abstraction that renders planning possible in the first place~\cite{konidaris_skills_2018}.

\begin{figure}[t]
  \centering
  \includegraphics[width=\linewidth]{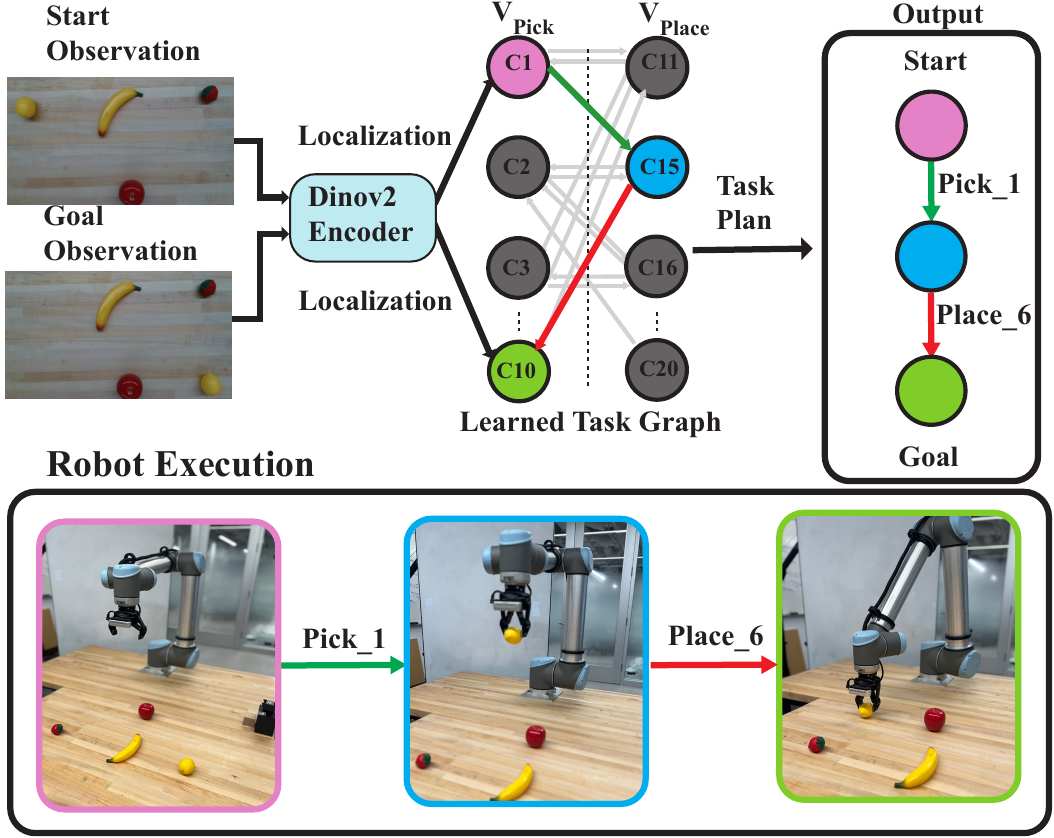}
\caption{\textbf{Image-To-Plan real-robot pipeline.}
We learn an action-labeled task graph from observed transitions $\mathcal{D}$.
At test time, given start and goal RGB observations (top-left), we localize each image to an abstract node in the learned graph.
We then plan between the localized nodes via discrete graph search (e.g., BFS) to obtain a high-level pick/place sequence, which the robot executes to transition the scene from start to goal (bottom).}
  \label{fig:fig1overview}
\end{figure}

We study this problem in the context of \emph{visual rearrangement tasks} (\autoref{fig:fig1overview}). Rearrangement tasks exhibit an underlying discrete structure and have become a canonical benchmark for task-planning research \cite{batra_rearrangement_2020,wang2022lazy}.

In our setting, neither the discrete states nor the transition dynamics are given: both must be inferred directly from image observations, without access to object identities, segmentation masks, geometric poses, or symbolic supervision. The robot is provided only with execution traces of the form $(o_t,a_t,o_{t+1})$, where $o_t$ is an RGB observation and $a_t \in \{\texttt{pick}_i,\texttt{place}_j\}$ is a high-level manipulation primitive. From this data, our objective is to infer an action-labeled abstract task graph  whose abstract nodes correspond to task-equivalence classes of observations and whose edges represent valid action transitions. This task graph can then be used to train a localization model that maps new observations to its nodes. Planning then reduces to graph search on the learned graph, followed by executing the resulting pick/place sequence on the robot (\autoref{fig:fig1overview}).

To learn such a graph, we explicitly leverage the universal structural constraints that govern real-world task dynamics. For example, all rearrangement tasks have bipartite task graphs, and between any ordered pair of abstract nodes, there is at most one action label. Violating these constraints compromises the soundness of the abstraction for planning. Prior works \cite{lippi_latent_2020, chamzas2022-contrastive-visual-task-planning}  learn task graphs through representation learning, relying primarily on visual similarity and auxiliary soft loss terms to encourage task-relevant clustering without explicitly enforcing structural feasibility of the resulting graph.

In contrast, we formulate the problem as a constrained graph construction problem. We explicitly encode multiple structural constraints and restrict the search to graphs that satisfy them by construction, using visual similarity only to guide selection among structurally valid candidates. We evaluate our method on several visual rearrangement tasks of varying difficulty, significantly outperforming pure representational learning methods in terms of task graph quality.
We also demonstrate the proposed method in two real-world visual rearrangement tasks, achieving image-to-plan success rates of 94\% and 70\%, respectively.

\section{Related Work}
Abstractions play a central role in enabling long-horizon robot planning by compressing high-dimensional sensorimotor data into compact, structured representations \cite{belta_symbolic_2007, konidaris_skills_2018}. We organize prior work along two axes: the form of abstraction learned, and the assumptions made about state representation.

\textbf{Relational and predicate-level abstraction:}
The foundational work of \cite{konidaris_skills_2018} learns STRIPS-style symbolic operators from skills, but assumes object-centric state descriptions and reliable action executability information. More recently,~\cite{silver_predicate_2025} extended this to learning expressive PDDL representations with first-order relational structure, again requiring structured object-level inputs and known operator semantics.   \cite{shah_real_2025, paulius_long-horizon_2023} tackle related long-horizon planning problems by learning generalizable relational concepts and functional object-oriented representations, respectively — both of which similarly rely on rich semantic or object-level structure. In contrast, \cite{asai_classical_2022} pioneered learning STRIPS-style abstractions directly from raw images, removing the need for predefined object models. Our approach shares this image-based motivation but additionally incorporates action-labeled transitions, which are often readily available in robotic settings.

\textbf{Monolithic transition-graph abstractions:}
A simpler and more direct form of abstraction is a flat transition graph \cite{bonet_blai_learning_2020}, where nodes represent discrete states and edges encode action transitions. Two main families of methods have been explored. The first jointly learns embeddings alongside forward or inverse dynamics models to encourage transition consistency in latent space \cite{bagatella_planning_2021, asai_classical_2022}.
The second follows a two-stage pipeline: learn a visual embedding and cluster observations to construct a graph, with edges defined via execution traces or latent distances \cite{chamzas2022-contrastive-visual-task-planning, lippi_latent_2020, yang_plan2vec_2020, zhang_composable_2018}. This is the closest work to ours and specifically the Latent Space Roadmap (\lsr)~\cite{lippi_latent_2020}, which constructs a visual task graph by clustering latent embeddings and connecting nodes via observed transitions. However, neither \lsr nor related approaches enforce structural constraints on the induced graph. As a result, learned graphs may contain structurally inconsistent edges. Our method addresses this directly by formulating abstraction learning as a constrained graph construction problem, restricting the search space to structurally valid graphs and using visual similarity only to select among feasible candidates.

\textbf{Visual rearrangement planning:}
Methods that focus specifically on visual rearrangement typically either establish dense image correspondences \cite{goyal2022ifor} or learn object-centric neural planners for unknown objects \cite{qureshi_nerp_2021}. Both approaches, however, require either object-level structure or correspondence supervision across views. In this work we address rearrangement under strictly weaker assumptions: the robot perceives the world only through raw RGB images and is provided with high-level pick-and-place action labels, with no access to object identities, poses, or segmentation masks. 

\section{Problem Formulation and Notation}
\label{sec:problem}

We begin by defining a task planning problem and the equivalent task-graph, and then specialize it to the case of rearrangement tasks, followed by the vision-based setting considered in this work.  

\textbf{Task Planning Problem and Task Graph} 
Let $\mathcal{S}$ be a finite set of discrete states, and let $\mathcal{A}$ be a finite set of discrete actions. The environment is governed by a deterministic transition function $T : \mathcal{S} \times \mathcal{A} \rightarrow \mathcal{S}$, which maps a state $s \in \mathcal{S}$ and an action $a \in \mathcal{A}$ to a successor state $s' = T(s,a)$.  
Given a start state $s_{\text{start}} \in \mathcal{S}$ and a goal state $s_{\text{goal}} \in \mathcal{S}$, the objective is to compute a \emph{plan}, i.e., a sequence of actions $\pi = (a_1, \dots, a_H)$ with $a_t \in \mathcal{A}$, such that executing $\pi$ transitions the system from $s_{\text{start}}$ to $s_{\text{goal}}$ under $T$.  

The task planning problem can be equivalently represented as a 
\emph{task graph} $G = (\mathcal{V}, \mathcal{E})$, where each 
node corresponds \emph{exactly} to a discrete state $s \in 
\mathcal{S}$\footnote{We distinguish nodes $\mathcal{V}$ from states $\mathcal{S}$ since states may have different representations (e.g., symbolic or factored), while nodes are discrete indexes}, and each directed labeled edge encodes a valid 
transition under $T$:
\begin{equation}
    \mathcal{E} = \{(s, a, s') \mid s, s' \in \mathcal{S},\ 
    a \in \mathcal{A},\ T(s, a) = s'\}.
\end{equation}
Note that the edge set $\mathcal{E}$ is richer than the action 
set $\mathcal{A}$: while $|\mathcal{A}|$ is fixed, the number 
of edges $|\mathcal{E}|$ grows with the state space, as the 
same action $a \in \mathcal{A}$ can induce transitions between 
many distinct state pairs. Given $G$, planning reduces to searching the graph for a labeled path from $s_{\text{start}}$ to $s_{\text{goal}}$.

\textbf{Rearrangement Planning Problem}
A rearrangement problem can be viewed as a special case of the task planning formulation with the additional structure:  
\begin{enumerate}
    \item The action set consists of pick and place actions, i.e., \mbox{$\mathcal{A} = \{\texttt{pick}_i, \texttt{place}_j\}$}, where the index $i$ distinguishes different pick/place actions\footnote{Depending on the task, pick actions could include pick\_red or pick\_left corresponding to picking a red object or picking from the left region.}. 
    \item The \emph{task graph} G is \emph{bipartite}: every $\texttt{pick}$ action must be followed by a $\texttt{place}$ (and vice-versa), and no two consecutive actions of the same type are valid.  
\end{enumerate}

%
%

\textbf{Visual Rearrangement Planning Problem}
In the visual rearrangement setting, the robot does not observe $\mathcal{S}$ directly. Instead, it perceives the world through raw visual observations $o \in \mathcal{O}$.  We assume an unknown observation map \mbox{$\phi : \mathcal{S} \rightarrow \mathcal{O}$} generates the visual observations from underlying states. In practice, multiple visually distinct observations may correspond to the same state due to sensor noise, continuous object placement or semantics. In our setting, the mapping $\phi$, the discrete state space $\mathcal{S}$, and the task-graph $G=(\mathcal{V},\mathcal{E})$ topology are all considered  unknown.

Instead, what is given is the set of discrete actions $\mathcal{A}$ and a dataset of observed execution traces:
\[
\mathcal{D} \;=\; \bigl\{(o_0,a_0,o_1),\ (o_1,a_1,o_2),\ \ldots,\ (o_{H-1},a_{H-1},o_H)\bigr\},
\]

where each $o_t \in \mathcal{O}$ is a visual observation and $a_t \in \{ \texttt{pick}_i, \texttt{place}_j \} $ is the corresponding action applied between successive observations.

The discrete action set $\mathcal{A}$ is assumed to be sufficiently expressive to realize all task-relevant transitions. For example, a task requiring all yellow objects on the right may include actions such as $\texttt{pick\_yellow}$ and $\texttt{place\_right}$ in $\mathcal{A}$. 
However, our method does not access these semantics; it observes only execution triples $(o_t,a_t,o_{t+1})$ and infers the abstract transition structure from the co-occurrence of observations and action labels. This assumption is consistent with the \emph{skills to symbols} perspective of \cite{konidaris_skills_2018}. 


Concretely, the \textbf{data-driven visual rearrangement planning 
problem} considered in this work is: Given the execution dataset 
$\mathcal{D}$, the discrete action set $\mathcal{A}$, a start 
observation $o_{\text{start}} = \phi(s_{\text{start}})$, and a 
goal observation $o_{\text{goal}} = \phi(s_{\text{goal}})$, find 
a sequence of actions $\pi = (a_1, \dots, a_H)$ that 
transitions the system from $s_{\text{start}}$ to $s_{\text{goal}}$.

\section{Approach} 
Our approach is to use the action set $\mathcal{A}$ and the dataset
$\mathcal{D}$ to construct a task graph
$\hat{\mathcal{G}} = (\hat{\mathcal{V}}, \hat{\mathcal{E}})$ that is
structurally similar to the unknown task graph $\mathcal{G}$.
We cast this graph construction problem as a clustering (or coloring)
problem over the observations $\mathcal{O}$.
Observations that are assigned to the same cluster (i.e., colored with
the same color) correspond to the same abstract node
$\hat{v} \in \hat{\mathcal{V}}$.
Once the observations are clustered, the edges of the graph follow
directly from the dataset: for each transition triple
$(o_t, a_t, o_{t+1}) \in \mathcal{D}$ we add an edge with the label of $a_t$
between the abstract nodes corresponding to the clusters containing
$o_t$ and $o_{t+1}$ yielding a complete task graph $\hat{G}$. A visual example of clustering observations to a graph through coloring is shown in \autoref{fig:exactlyone}. 

To use this graph for planning with novel start/goal observations we train an image-to-node classifier  $\hat{f}_{\theta}:\mathcal{O}\to \hat{\mathcal{V}}$ using the induced cluster
labels as supervision. Planning then reduces to graph search (e.g., BFS) on
$\hat{\mathcal{G}}$ between $\hat{f}_{\theta}(o_{\text{start}})$ and
$\hat{f}_{\theta}(o_{\text{goal}})$.

\subsection{Structural Task Constraints}
\label{sec:overall}

We enforce two structural constraints that arise naturally in
rearrangement tasks.

\textbf{Bipartiteness.}
Rearrangement task graphs are bipartite: pick actions must be
followed by place actions and vice versa, so no pick$\!\to\!$pick or
place$\!\to\!$place transitions can occur.
Accordingly, the learned graph is constrained to have the form
$
\hat{\mathcal{G}} =
(\hat{\mathcal{V}}_{\text{pick}} \cup \hat{\mathcal{V}}_{\text{place}},
\hat{\mathcal{E}}),
$
where $\hat{\mathcal{V}}_{\text{pick}}$ are abstract nodes from which pick actions can be executed and $\hat{\mathcal{V}}_{\text{place}}$ abstract nodes from which place actions can be executed.

\begin{figure}[t]
  \centering
  \includegraphics[width=0.7\columnwidth]{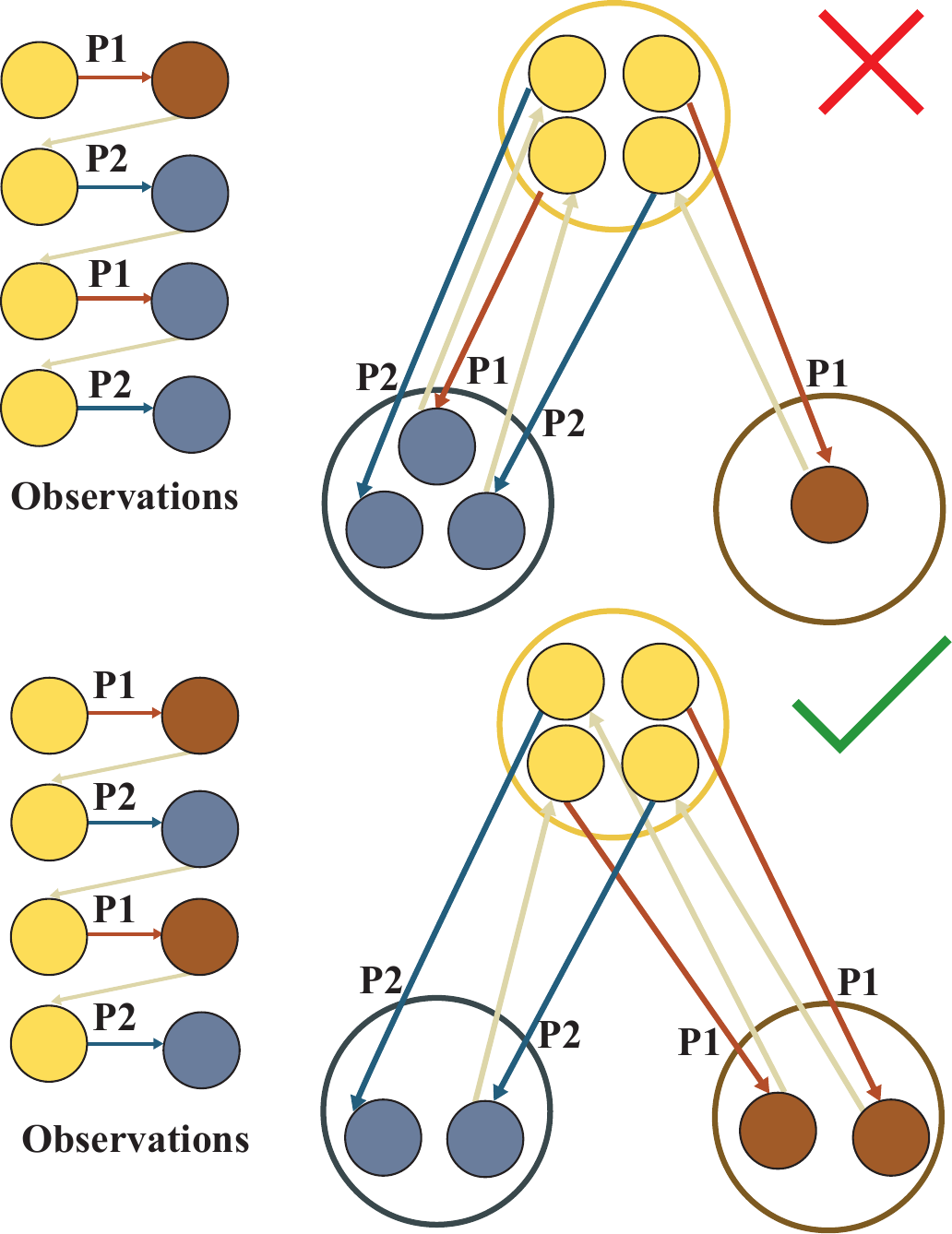}
\caption{\textbf{Observations coloring with the action–uniqueness constraint:}
Each small circle corresponds to an observation. Coloring the observations with the same color means that they belong to the same abstract node (same cluster).  
\textcolor{red}{Top:} \emph{invalid}—between the same colored clusters we observe two different actions (e.g., $p_1$ and $p_2$); this violates our rule that there must be a unique action between any fixed pair of abstract states.
\textcolor{green!60!black}{Bottom:} \emph{valid}—only one action connects the pair; any second action must lead to a different destination cluster.}

  \label{fig:exactlyone}
\end{figure}

\textbf{Action-uniqueness (Exactly-One).} Between any ordered pair of abstract states, there is at most one action label. This constraint is shown in \autoref{fig:exactlyone}:

\begin{equation}
\forall v, v'\in\hat{\mathcal{V}}:\quad
\bigl|\{\,a\in\mathcal{A} : (v,a,v')\in\hat{\mathcal{E}}\,\}\bigr|\ \le\ 1.
\label{eq:exact1}
\end{equation}

These constraints rule out structurally inconsistent task-planning graphs, but they still leave many feasible ways to group observations into nodes. We therefore use a vision-based distance (\autoref{sec:visual-sim}) to score feasible groupings. 

\label{sec:exactly-one}

\subsection{Attention-Guided Visual Distance}
\label{sec:visual-sim}

We aim to find  a distance that is \emph{small} when two images have objects in similar locations, and large when they differ in the spatial arrangement that matters for pick-and-place.

\subsubsection{\textbf{Attention maps}}
We compute a soft spatial distribution $M(o)$ for each RGB observation $o$ using a frozen Vision Transformer\cite{oquab2024dinov2}. We extract the final-layer patch attention weights
and normalize them over the patch grid so that $\sum_{ij} M_{ij}(o)=1$. This gives us an object-agnostic weighting of image regions that the backbone considers salient.

We compare two attention maps by treating them as distributions over patch locations and computing a debiased, entropically regularized optimal-transport cost~\mbox{\cite{cuturi2013sinkhorn, feydy2019sinkhorn}}(Sinkhorn divergence). Let $p$ and $q$ be the vectorized maps for observations $o_i$ and $o_j$, and let the ground cost between patches $u,v$ be the Euclidean distance between their centers, $C_{uv}=\|x_u-x_v\|_2$.

Let $\mathrm{OT}_{\varepsilon}(p,q)$ denote the entropically regularized OT cost with regularization $\varepsilon$.
We use the debiased Sinkhorn divergence
\begin{equation}
S_{\varepsilon}(p,q)=\mathrm{OT}_{\varepsilon}(p,q)
-\tfrac{1}{2}\mathrm{OT}_{\varepsilon}(p,p)
-\tfrac{1}{2}\mathrm{OT}_{\varepsilon}(q,q),
\label{eq:sinkhorn-div}
\end{equation}
and define the visual distance
\begin{equation}
d_{\mathrm{vis}}(o_i,o_j)\;=\;S_{\varepsilon}\!\bigl(M(o_i),M(o_j)\bigr).
\label{eq:visdist}
\end{equation}

We convert distances to a visual affinity
$K_{\mathrm{vis}}(o_i,o_j)=\exp(-d_{\mathrm{vis}}(o_i,o_j)/\tau)$.
Given a partition $\mathcal{C}=\{C_1,\dots,C_K\}$ of a role-specific observation set (i.e., a clustering into $K$ abstract nodes), we score its visual coherence by the mean intra-cluster distance
\begin{equation}
\Vis(\mathcal{C})=\sum_{C\in\mathcal{C}}
\frac{1}{|C|(|C|-1)}
\sum_{\substack{o,o'\in C\\o\neq o'}} d_{\mathrm{vis}}(o,o').
\label{eq:vis-score}
\end{equation}

Lower $\Vis(\mathcal{C})$ indicates that observations mapped to the same abstract node are visually consistent in task-relevant image regions.

\section{Methodology}

\label{sec:method}
\begin{figure*}[t]
  \centering
  \includegraphics[width=\textwidth]{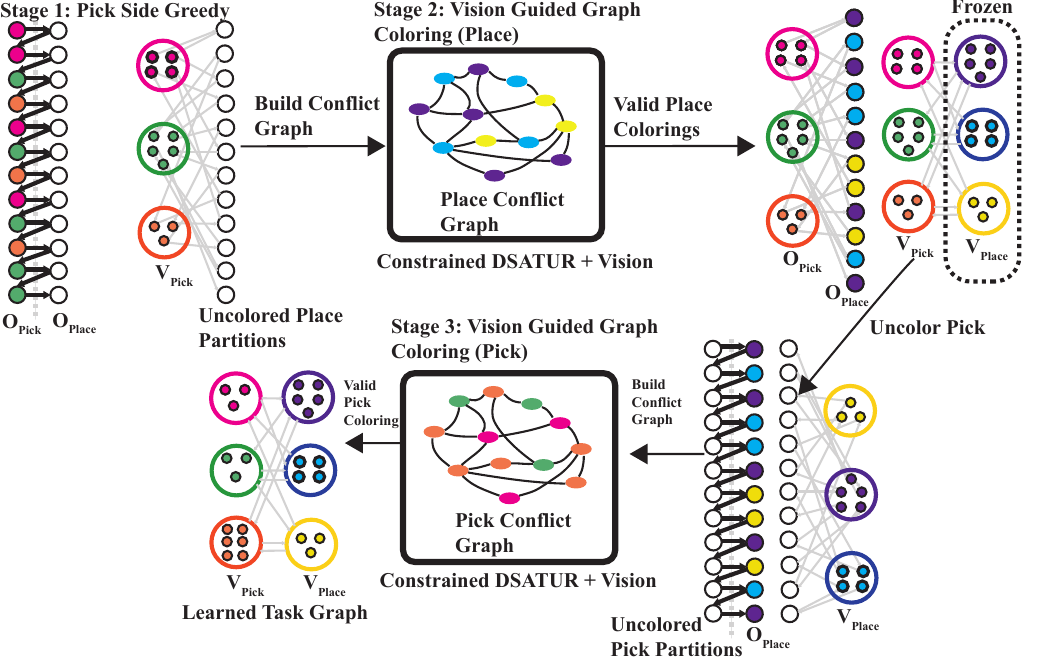}
\caption{\textbf{Vision–guided graph coloring.}
\emph{Stage 1} (top-left): seed the pick side by greedy, vision‑only grouping
using $K_{\mathrm{vis}}$ (no relational checks).
\emph{Stage 2} (top–middle): build the \emph{place conflict graph} whose
edges indicate merges that would violate Exactly‑One;
color with DSATUR, choosing among admissible colors by visual affinity.
\emph{Stage 3} (bottom–middle): freeze place and recolor the pick side the
same way; the bipartite graph now satisfies the constraints.
}
 
  \label{fig:graphcoloring}
\end{figure*}

\subsection{Vision-Guided Graph Coloring}
\label{sec:coloring-intuition}

Let $\mathcal{O}^{\text{pick}}$ and $\mathcal{O}^{\text{place}}$ denote the observations that appear immediately before pick and place actions in $\mathcal{D}$, respectively (the role is known from the action stream).
A naive approach would enumerate all partitions of $\mathcal{O}^{\text{pick}}$ into $k_{\text{pick}}$ clusters and all partitions of $\mathcal{O}^{\text{place}}$ into $k_{\text{place}}$ clusters, filter those that satisfy the structural constraints, and then select the visually most coherent solution.
This is intractable because the number of set partitions grows super-exponentially (Bell numbers).
Instead, we directly solve a constraint-guided coloring problem that enforces feasibility during assignment. Since $(k_{\text{pick}},k_{\text{place}})$ are unknown, we evaluate a small grid and select the best feasible graph using \Vis\ (\autoref{sec:graph-selection}).
Because pick and place are colored separately, the learned bipartite graph contains $K = k_{\text{pick}} + k_{\text{place}}$ abstract nodes in total.

We now describe how we compute this partition efficiently while enforcing the constraints.

\paragraph*{\textbf{Stage 1: Vision‑only seeding of the pick side} (\autoref{fig:graphcoloring}, top-left)}
The Exactly-One constraint couples pick and place assignments: deciding whether two \emph{place} observations can share a cluster depends on which \emph{pick} cluster ($\mathcal{V}_{\text{pick}}$) they transition from (and vice versa). A pick cluster is a cluster from which a pick action can be executed, and a place cluster ($\mathcal{V}_{\text{place}}$) is where a place action can be executed. We therefore begin by constructing a tentative partition of the pick observations, which we treat as fixed context when coloring the place side in Stage~2.

We build these seeds using only pairwise \emph{visual affinity} $K_{\mathrm{vis}}$ (\autoref{eq:visdist}), which measures similarity between two observations; the overall \emph{visual coherence} of a full partition is quantified by \Vis\ (\autoref{eq:vis-score}). Concretely, we select an unassigned pick observation with high mean affinity to the rest as a seed, start a new color, and greedily add the remaining pick observations that minimize $d_{\mathrm{vis}}$ while keeping the cluster's mean distance low. This generates compact initial clusters without yet checking relational constraints. The number of colors is bounded by $k_{\text{pick}}$.

\paragraph*{\textbf{Stage 2: Place‑side DSATUR under constraints} (\autoref{fig:graphcoloring}, top-middle)}
Holding the Stage‑1 pick partition fixed, we build the \textbf{place conflict graph}: two place images are adjacent if putting them in the same color would violate the Exactly-One constraint \eqref{eq:exact1} when combined with the \emph{already colored} pick partition and the observed transitions $(o_t,a_t,o_{t+1})$ in $\mathcal{D}$. We then color this graph with at most $k_{\text{place}}$ colors using DSATUR~\cite{brelaz1979dsatur}.
\begin{itemize}\itemsep2pt
  \item \textbf{Ordering.} At each step we select the place image with the highest saturation degree (most distinct colors seen among its neighbors), breaking ties by total degree.
  \item \textbf{Vision‑guided color choice.} Among \emph{feasible} colors, we choose the one with the highest mean $K_{\mathrm{vis}}$ affinity to its current members, thereby integrating vision into the combinatorial step.
\end{itemize}

\paragraph*{\textbf{Stage 3: Pick‑side DSATUR with place frozen} (\autoref{fig:graphcoloring}, bottom–middle)}
After obtaining a feasible coloring for the \textit{place} side in Stage~2, we \emph{discard the greedy seeds on the pick side} from Stage~1 (i.e., uncolor all pick nodes) and \emph{recolor pick from scratch} while holding the place partition fixed.
Concretely, we build the \textbf{pick conflict graph}: two pick images are adjacent if merging them into the same cluster would violate the Exactly-One constraint \autoref{eq:exact1} when combined with the \emph{frozen} place partition and the observed transitions $(o_t,a_t,o_{t+1})\in\mathcal{D}$.
We then color this graph with at most $k_{\text{pick}}$ colors using DSATUR, using the same policy as in Stage~2:
After this stage, both sides have been colored against each other and the resulting action‑labeled bipartite graph satisfies the constraints by construction.

\paragraph*{\textbf{Feasibility and backtracking}}
When testing an assignment $v\!\rightarrow\!c$ during DSATUR (Stages 2–3), we maintain a \emph{pair–action table} $\Lambda$ that records, for each ordered pair of clusters $(c,c')$, the unique action label $\Lambda(c,c')$ already fixed by earlier decisions. An assignment is \emph{feasible} iff:
\begin{enumerate}\itemsep2pt
  \item \textbf{No neighbor conflict.} No neighbor of $v$ in the current role’s conflict graph already has color $c$.
 \item \textbf{Exactly-One.} For every observed transition $(v,a,v')$ whose opposite-role partner $v'$ is already colored with color $c'$, either $\Lambda(c,c')$ is unset or equals $a$.
\end{enumerate}
If several colors are feasible we use the vision‑guided rule above. If no existing color is feasible and we have not yet used $k_{\text{pick}}$/$k_{\text{place}}$ colors, we \emph{open} a new color. Otherwise we \emph{backtrack} to the most recent undecided vertex and try its next feasible color. In practice, DSATUR’s “most‑constrained‑first’’ ordering keeps backtracks shallow.

\subsection{Learned Graph Selection}
\label{sec:graph-selection}
\begin{figure}[t]
  \centering
  \includegraphics[width=0.85\columnwidth]{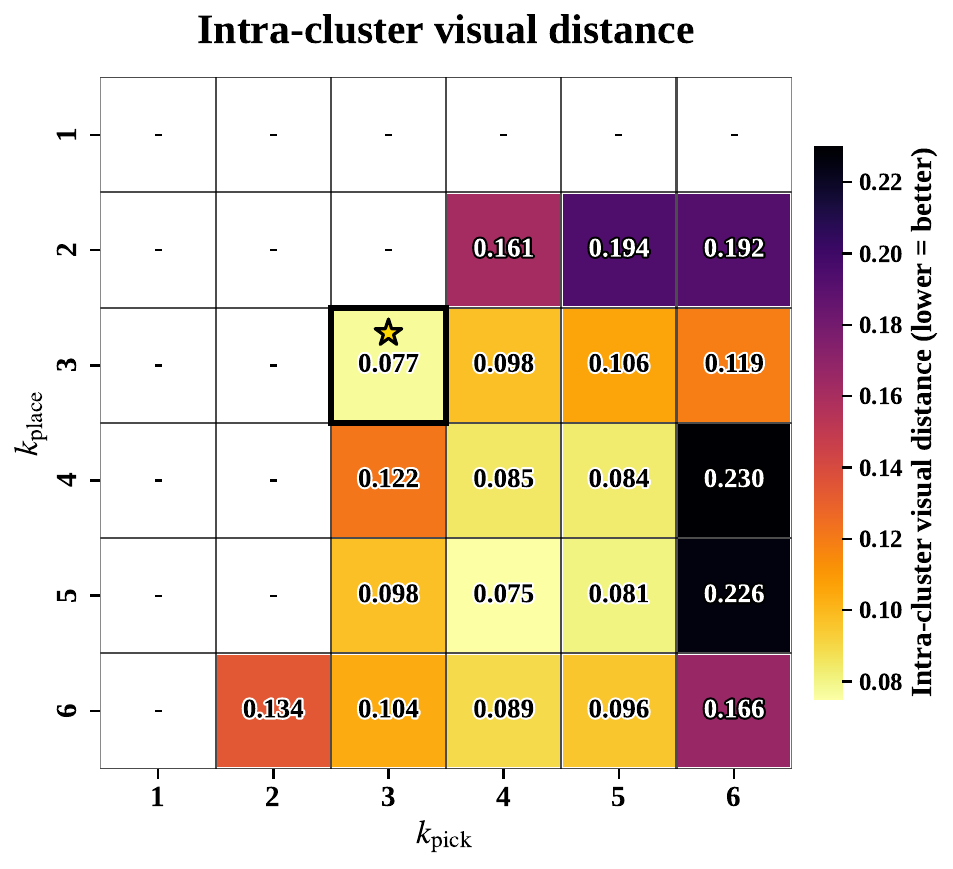}
\caption{\textbf{Grid sweep.} Evaluating $(k_{\text{pick}},k_{\text{place}})$.
Lower \Vis\ is better. For Fruit-2$\times$3, the best is $(3,3)$. Blank cells indicate no solution was found for that pair.}
\label{fig:sweep}

\end{figure}

\textbf{Selection across $(k_{\text{pick}},k_{\text{place}})$:}
We repeat the three‑stage procedure above across a small grid of $(k_{\text{pick}},k_{\text{place}})$. For each grid cell we keep the best feasible partition (lowest intra-cluster visual distance $\Vis$) and then select the final abstraction by (i) fewest total clusters $k_{\text{pick}}{+}k_{\text{place}}$, and (ii) minimum \Vis to break ties. The resulting sweep is summarized in \autoref{fig:sweep}; the selected graph is the one we use for planning and evaluation.

\subsection{Few‑Shot Visual–to–State Classification}
\label{sec:fewshot}
To localize any \emph{new} observation $o_{\text{new}}$ on the learned task graph, we must map a new observation $o_{\text{new}}$
to a node of the learned task-planning graph $\hat{\mathcal{G}}$.
We therefore train an image-to-node classifier $\hat{f}_\theta:\mathcal{O}\rightarrow\hat{\mathcal{V}}$
that generalizes the discrete assignment produced during graph learning. 

\noindent\textbf{Labels from the learned graph.}
The learned partition provides supervision at no additional annotation cost.
Let $K = |\hat{\mathcal{V}}|$ be the number of learned nodes and fix an index map
$\mathrm{idx}:\hat{\mathcal{V}}\to\{1,\dots,K\}$.
For each training image $o_i$ we assign the label
$y_i := \mathrm{idx}(f(o_i))$, yielding the classification set
$\mathcal{D}_{\mathrm{cls}}=\{(o_i,y_i)\}$.

\noindent\textbf{Classifier.}
We use a pretrained DINOv2\cite{oquab2024dinov2} ViT-G/14 encoder to extract features $\Phi(o)\in\mathbb{R}^d$,
and attach a cosine-similarity classifier head:
\begin{equation}
z_k(o)\;=\;\tfrac{1}{\gamma}\,
\frac{w_k^\top \Phi(o)}{\|w_k\|\,\|\Phi(o)\|},
\qquad
\hat{y}(o)\;=\;\arg\max_k z_k(o),
\label{eq:cosine-head}
\end{equation}
where $w_k\in\mathbb{R}^d$ are trainable class weights and $\gamma$ is a temperature.

\noindent\textbf{Training in the low-data regime.}
Because each node has few examples, we train with a lightweight fine-tuning setup:
we keep the backbone largely frozen (fine-tuning only the final transformer block)
and optimize the cosine head with cross-entropy.
We use light image augmentations to improve robustness to appearance changes and
perform simple cross-validation on $\mathcal{D}_{\mathrm{cls}}$ to select
training hyperparameters. A single forward pass maps $o_{\text{new}}$ to a node in $\hat{\mathcal{G}}$.

\begin{table}[t]
\centering
\setlength{\tabcolsep}{4pt}
\renewcommand{\arraystretch}{1.12}
\scriptsize
\begin{tabular*}{\columnwidth}{@{\extracolsep{\fill}} lcccccccc @{}}
\toprule
\textbf{Benchmark} & \textbf{Sim} & \textbf{Real} &
$\mathbf{m}$ & $\mathbf{n}$ & $\mathbf{R}$ & $\mathbf{|\mathcal{A}|}$ &
$\mathbf{|\mathcal{V}^\star|}$ &
$\mathbf{|\mathcal{D}|}$ \\
\midrule
Fruit-2$\times$3   & \checkmark & \checkmark & 2 & 1 & 3 & 6  & 6   & \;\;\;50 \\
Multi Fruit-2$\times$3   & \checkmark &            & 2 & 2 & 3 & 12 & 12  & \;\;\;95 \\
Fruit-4$\times$6        & \checkmark &            & 4 & 1 & 6 & 12 & 35  & \;\;\;700 \\
Multi Fruit-4$\times$6  & \checkmark & \checkmark & 4 & 2 & 6 & 24 & 58  & \;\;\;400 \\
Fruit-6$\times$8        & \checkmark &            & 6 & 1 & 8 & 16 & 84  & \;\;\;1500 \\
Blocks-2                & \checkmark &            & 2 & 1 & 3 & 12 & 9   & \;\;\;95 \\
Blocks-3                & \checkmark &            & 3 & 1 & 3 & 18 & 16  & \;\;\;200 \\
\bottomrule
\end{tabular*}

\caption{\textbf{Benchmark suite.} $m$ = number of objects, $n$ = number of task-distinct objects, $R$ = number of regions,
$|\mathcal{A}|$ = number of discrete action labels, $|\mathcal{V}^\star|$ = number of nodes in the ground-truth graph $G^\star$ and $|\mathcal{D}|$ = number of observed transitions used to generate the learned graph.}
\label{tab:benchmarks}
\end{table}



\label{sec:ablation}

\section{Experimental Setup}
\label{sec:experiments}\label{sec:overview}
We evaluate on tabletop \emph{visual rearrangement} tasks of increasing combinatorial difficulty:
(i) \textbf{Table-Top Rearrangement} (\textsc{Fruit}) and
(ii) \textbf{Stacking Rearrangement} (\textsc{Blocks World}).
All experiments use only raw RGB images and discrete pick/place action labels; we do not assume object identities,
poses, or segmentation masks.
We evaluate (i) whether the learned abstract graph supports correct high-level planning and (ii) whether the induced graph is structurally valid and visually coherent.

\subsection{Simulation setup}
We use \textbf{PyBullet}~\cite{coumans2016pybullet} with a fixed RGB camera. Objects are taken from the \textbf{YCB object set}~\cite{ycb}; we use a subset consisting primarily of common fruits and blocks for both our experiments in simulation and real hardware.

For each benchmark, we collect an exploration trace by executing random valid pick/place actions (respecting pick--place alternation and region occupancy constraints) and recording the resulting RGB observation after each action, yielding a dataset of transitions $\mathcal{D}=\{(o_t,a_t,o_{t+1})\}$.
\autoref{tab:benchmarks} reports the number of transitions $|\mathcal{D}|$ used to construct each learned graph. 


\subsection{Tasks}
\label{sec:tasks}
We evaluate the benchmark suite summarized in \autoref{tab:benchmarks}. 
For each benchmark, the methods are provided with the discrete action set $\mathcal{A}$ and an execution trace $\mathcal{D}$. As discussed earlier, $\mathcal{A}$ is assumed to be sufficiently expressive to realize all task-relevant transitions.

Across all rearrangement tasks, the scene contains $m$ physical objects placed among $R$ non-overlapping regions. Depending on the benchmark, only $n$ of these objects are \emph{task-distinct} (with $n \le m$)\footnote{Task-distinct objects are treated as different by the task specification (e.g., different categories), while the remaining objects are interchangeable.}. 
Accordingly, the action set contains $nR$ distinct pick actions and $nR$ distinct place actions, for a total of $|\mathcal{A}| = 2nR$ discrete actions.

\subsubsection{\textsc{Fruit}: tabletop rearrangement}
\textsc{Fruit} consists of $m$ objects distributed among $R$ tabletop regions, with $m<R$ so that at least one region is empty.
For multi-object variants, we use $n{=}2$ visual classes:
\textsc{Multi-Fruit}~2$\times$3 includes two distinct fruits (strawberry and lemon), while \textsc{Multi-Fruit}~4$\times$6 includes four unique objects (strawberry, apple, lemon, banana) spanning two visual classes (\texttt{red} and \texttt{yellow}).

\subsubsection{\textsc{Blocks World}: stacking}
\textsc{Blocks World} (inspired by~\cite{lippi_latent_2020}) uses $R{=}3$ base regions (left/center/right), each of which can hold a stack.
Blocks are treated as interchangeable.
\textsc{BlocksWorld-2} uses 2 blocks with maximum stack height 2, and \textsc{BlocksWorld-3} uses 3 blocks with maximum stack height 3.

\subsubsection{Real-robot tasks}
On hardware we mirror the \textsc{Fruit} setup and report image-to-plan results from RGB start/goal images using a UR10 and an Intel RealSense D435.
We evaluate Fruit-2$\times$3 and Multi Fruit-4$\times$6 (\autoref{tab:benchmarks}).
\subsection{Scalability Tests}
\label{sec:scalability_tests}

We study how performance and runtime scale with the amount of available data.
We fix the underlying environment and action interface (we use \textbf{Fruit-4$\times$6}) and vary the number of
observed transitions $|\mathcal{D}|$ used to learn the task-planning graph, with
$|\mathcal{D}|\in\{30,70,100,200,500,1000,1500,2000\}$.
For each dataset size $|\mathcal{D}|$, we run the full pipeline on the corresponding dataset and report (i) planning quality on the learned graph
(using the metrics in (\autoref{sec:metrics}) and (ii) wall-clock runtime (and, when applicable, a breakdown by major
stages such as distance computation and search time).

\newcommand{\lsrbase}{\makecell[l]{LSR\cite{lippi_latent_2020}}}
\newcommand{\hdbbase}{\makecell[l]{PCA--HDB-RM}}
\newcommand{\mOpt}{\makecell{Opt\\$\uparrow$}}
\newcommand{\mAny}{\makecell{Any\\$\uparrow$}}
\newcommand{\mTP}{\makecell{Trans\\Prec\\$\uparrow$}}
\newcommand{\mCov}{\makecell{Trans\\Cov\\$\uparrow$}}

\newcommand{\best}[1]{\textbf{#1}}     
\newcommand{\bestv}[1]{\textbf{#1}}    

\begin{table*}[t]
\centering
\setlength{\tabcolsep}{2.6pt}
\renewcommand{\arraystretch}{1.05}
\scriptsize

\begin{tabularx}{\linewidth}{@{}>{\raggedright\arraybackslash}p{1.75cm} *{16}{>{\centering\arraybackslash}X}@{}}
\toprule
& \multicolumn{4}{c}{Fruit-2$\times$3}
& \multicolumn{4}{c}{Multi Fruit-2$\times$3}
& \multicolumn{4}{c}{Blocks-2}
& \multicolumn{4}{c}{Blocks-3} \\
\cmidrule(lr){2-5}\cmidrule(lr){6-9}\cmidrule(lr){10-13}\cmidrule(lr){14-17}
\textbf{Method}
& \mOpt & \mAny & \mTP & \mCov
& \mOpt & \mAny & \mTP & \mCov
& \mOpt & \mAny & \mTP & \mCov
& \mOpt & \mAny & \mTP & \mCov \\
\midrule
\lsrbase
& 100\% & 100\% & 100\% & 100\%
& 40\% & 70\% & 45.9\% & 77.3\%
& 76.7\% & 76.7\% & \best{100\%} & 83.0\%
& 33.6\% & 60.7\% & 57.1\% & 69.4\% \\
\hdbbase
& 100\% & 100\% & 100\% & 100\%
& 32.0\% & 32.0\% & 55.3\% & 63.6\%
& 16.1\% & 16.1\% & 42.9\% & 33.3\%
& 30.8\% & 32.5\% & 25.0\% & 40.5\% \\
\textbf{VGGC (Ours)}
& \best{100\%} & \best{100\%} & \best{100\%} & \best{100\%}
& \best{66.0\%} & \best{100\%} & \best{83.3\%} & \best{91.7\%}
& \best{93.7\%} & \best{100\%} & \best{100\%} & \best{90\%}
& \textbf{72.0\%} & \textbf{100\%} & \textbf{77.8\%} & \textbf{67.7\%} \\
\bottomrule
\end{tabularx}

\begin{tabularx}{\linewidth}{@{}>{\raggedright\arraybackslash}p{1.75cm} *{12}{>{\centering\arraybackslash}X}@{}}
\toprule
& \multicolumn{4}{c}{Fruit-4$\times$6}
& \multicolumn{4}{c}{Multi Fruit-4$\times$6}
& \multicolumn{4}{c}{Fruit-6$\times$8} \\
\cmidrule(lr){2-5}\cmidrule(lr){6-9}\cmidrule(lr){10-13}
\textbf{Method}
& \mOpt & \mAny & \mTP & \mCov
& \mOpt & \mAny & \mTP & \mCov
& \mOpt & \mAny & \mTP & \mCov \\
\midrule
\lsrbase
& 37.5\%  & 61.7\%  & 50.0\%  & 53.3\%
& 12.0\% & 12.0\% & 23.5\% & 49.2\%
& 52.7\%  & 88.9\%  & 51.2\%  & 65.2\% \\
\hdbbase
& 33.4\%  & 56.9\%  & 58.6\%  & 56.7\%
& 36.6\%  & 88.9\%  & 60.6\%  & 57.4\%
& 52.1\% & 84.1\% & 44.0\% & 56.9\% \\
\textbf{VGGC (Ours)}
& \best{78.1\%} & \best{99.1\%} & \best{81.6\%}   & \best{85.8\%}
& \best{71.8\%} & \best{98.2\%} & \best{88.5\%} & \best{81.6\%}
& \best{64.8\%} & \best{95.7\%} & \best{72.9\%} & \best{84.7\%} \\
\bottomrule
\end{tabularx}
\caption{\textbf{Planning and graph-quality results.}
Opt/Any measure shortest-plan and any-plan validity against $\mathcal{G}^\star$.
TransPrec and TransCov report edge precision/coverage against $\mathcal{G}^\star$.
\textbf{PCA--HDB-RM} denotes the PCA--HDBSCAN Roadmap baseline~\cite{mcinnes2017hdbscan}. Reported over Q=1500 different start and goal nodes.}
\label{tab:main_results}
\end{table*}

\begin{table}[t]
\centering
\setlength{\tabcolsep}{3.6pt}
\renewcommand{\arraystretch}{1.08}
\scriptsize

\begin{tabularx}{\columnwidth}{@{}l c *{5}{>{\centering\arraybackslash}X} @{}}
\toprule
\textbf{Env (VGGC)} & \textbf{\#Cls} &
\makecell{\textbf{Opt}\\$\uparrow$} &
\makecell{\textbf{Any}\\$\uparrow$} &
\makecell{\textbf{TransCov}\\$\uparrow$} &
\makecell{\textbf{Trans}\\\textbf{Prec}\\$\uparrow$} &
\makecell{\textbf{I2P}\\$\uparrow$} \\
\midrule
Fruit-2$\times$3       & 6& 94.7\%  & 100\%  &  85.7\% & 100\%  &    94\%    \\
Multi Fruit-4$\times$6 & 58 & 46.7\% & 94.7\% & 76.7\% & 69.2\% &  70\%     \\
\bottomrule
\end{tabularx}

\caption{\textbf{Real-robot evaluation (VGGC).} Reported over $Q{=}150$ different start and goal images.}
\label{tab:real_robot}
\end{table}

\subsection{Ground-truth graph for evaluation ($G^\star$)}
In simulation we have access to the underlying discrete scene configuration for each observation
(e.g., region occupancy and, in \textsc{Blocks World}, stack order).
We use this configuration as the ground-truth state $v \in \mathcal{V}^\star$ and construct a ground-truth transition graph
$G^\star=(\mathcal{V}^\star,\mathcal{E}^\star)$ from the action-labeled transitions in the dataset.
We emphasize that $G^\star$ is used \emph{only} for evaluation; our method never uses ground-truth states or transitions for learning.


\subsubsection{Plan validation in $G^\star$}
Let $o_{\text{start}}$ and $o_{\text{goal}}$ denote the start and goal observations, and let
$\hat{v}_{\text{start}}$=$\hat{f}_{\theta}(o_{\text{start}})$ and
$\hat{v}_{\text{goal}}=\hat{f}_{\theta}(o_{\text{goal}})$
be their localized abstract nodes.

We say a ground-truth state $v\in\mathcal{V}^\star$ is \emph{consistent} with an abstract node $\hat{v}$ if the dataset contains at least one observation $o$
with underlying state $v$ such that $f(o)=\hat{v}$.
An abstract action sequence $\pi=(a_1,\dots,a_L)$ is deemed valid if there exists a corresponding sequence of ground-truth states $(v_0,\dots,v_L)$ such that
(i) $v_0$ is consistent with $\hat{v}_{\text{start}}$,
(ii) each transition $(v_{t},a_{t+1},v_{t+1})\in \mathcal{E}^\star$ for all $t\in\{0,\dots,L{-}1\}$, and
(iii) $v_L$ is consistent with $\hat{v}_{\text{goal}}$.
If any step is invalid, the plan is counted as failure.



\subsection{Planning metrics}
\noindent\textbf{OptPath (\%)}: fraction of plans for which at least one \emph{shortest-length} abstract plan in $\hat{G}$ validates in $G^\star$.\\
\textbf{AnyPath (\%)}: fraction of plans for which \emph{some} abstract plan (not necessarily shortest) validates in $G^\star$.

\subsubsection{Graph quality metrics}
\label{sec:metrics}
Clustering can alias multiple ground-truth states into a single abstract node; under aliasing, an abstract edge may be correct for some underlying states and incorrect for others.
We therefore evaluate learned edges against $G^\star$ using:\\
\noindent\textbf{TransPrec (\%)}: fraction of directed labeled edges $(\hat{v},a,\hat{v}')\in \hat{\mathcal{E}}$ such that there exist ground-truth states $v,v'\in\mathcal{V}^\star$ with $v$ consistent with $\hat{v}$, $v'$ consistent with $\hat{v}'$, and $(v,a,v')\in \mathcal{E}^\star$.\\
\textbf{TransCov (\%)}: fraction of ground-truth edges $(v,a,v')\in \mathcal{E}^\star$ such that there exist abstract nodes $\hat{v},\hat{v}'$ with $v$ consistent with $\hat{v}$, $v'$ consistent with $\hat{v}'$, and $(\hat{v},a,\hat{v}')\in \hat{\mathcal{E}}$.\\
Intuitively, TransPrec measures how “clean” the learned edges are (precision), while TransCov measures how much of the true dynamics is captured (recall).

\subsection{Baselines}
We compare our method against two baselines that operate directly on RGB observations and discrete action labels, and do not require object-centric inputs such as segmentation masks.
First, we evaluate a \textbf{PCA--HDBSCAN Roadmap} baseline~\cite{mcinnes2017hdbscan} that applies PCA to the RGB observations, clusters $\mathcal{O}^{\text{pick}}$ and $\mathcal{O}^{\text{place}}$ separately with HDBSCAN, and constructs an action-labeled abstract graph by aggregating the observed transitions between clusters.
Second, we evaluate \textbf{Latent Space Roadmaps (LSR)}~\cite{lippi_latent_2020}, which clusters observations in a learned embedding space and constructs a roadmap over the resulting clusters.
For a fair comparison, we pre-split the dataset into $\mathcal{O}^{\text{pick}}$ and $\mathcal{O}^{\text{place}}$, run each baseline independently on the two partitions, and sweep hyperparameters for both methods.

\subsubsection{Image-to-Plan task success (I2P)}
To measure image-to-plan performance, integrating perception (few-shot localization) with planning on the learned graph, we report \textbf{I2P (\%)} (Image-to-Plan success rate). This metric combines \emph{perception + planning} success from real RGB images; we assume the underlying pick/place primitives execute reliably and do not attribute failures to low-level control.

Given start/goal RGB images, we first localize each image to an abstract node in the learned task graph $\hat{G}$ using the few-shot classifier.
If either image is misclassified, the trial is counted as a failure.
Otherwise, we plan on $\hat{G}$ between the localized nodes and evaluate success by validating the plan against the ground-truth graph $G^\star$.
For real-robot trials, $G^\star$ is constructed from a matched simulation trace for the same benchmark.
A trial is counted as successful iff at least one plan found in $\hat{G}$ is valid in $G^\star$.

\section{Results}
\begin{figure}[t]
  \centering
  \includegraphics[width=0.80\columnwidth]{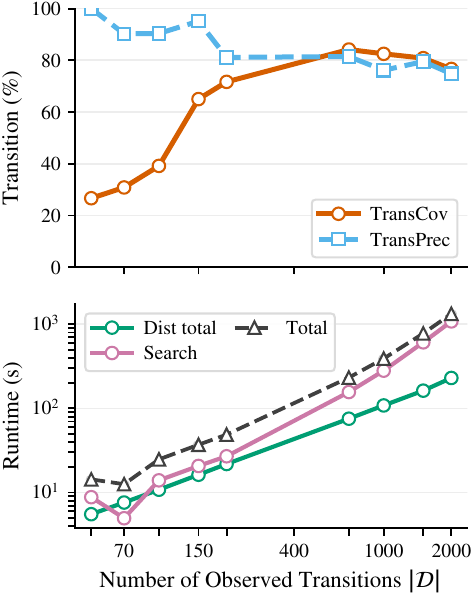}
  \caption{\textbf{Scalability on Fruit-4$\times$6.}
  Top: TransCov/TransPrec vs.\ $G^\star$ as $|\mathcal{D}|$ increases.
  Bottom: runtime breakdown (distance vs.\ search).}
  \label{fig:scalability}
\end{figure}

We compare \textbf{VGGC} against two baselines in simulation and on  real-robot benchmarks.
\autoref{tab:main_results} summarizes planning success (Opt/Any) and graph fidelity (TransPrec/TransCov) across the full benchmark suite.
Overall, \textbf{VGGC outperforms the baselines on 6 of 7 benchmarks}, with the largest gains appearing in the harder settings with more observations and larger action alphabets.
On the simplest benchmark (Fruit-2$\times$3), all methods achieve perfect performance, indicating that differences emerge primarily when abstraction learning is non-trivial. 
As task complexity increases, the baselines degrade, while VGGC remains robust.
In particular, VGGC consistently produces \emph{cleaner} graphs, as measured by higher transition precision, while also maintaining high transition coverage on the larger benchmarks. 
For example, on Fruit-6$\times$8 ($|\mathcal{D}|$ =1500), VGGC achieves $>70\%$ TransPrec, substantially exceeding both baselines. Overall, the gains are largest when naive clustering creates structurally inconsistent merges; enforcing bipartiteness  + action-uniqueness during clustering prevents these merges, improving global connectivity.

We study scalability by fixing the environment to Fruit-4$\times$6 and increasing  $|\mathcal{D}|$ up to $2000$.
\autoref{fig:scalability} shows that \textbf{TransCov increases rapidly and then plateaus} as $|\mathcal{D}|$ grows, suggesting that additional data primarily fills in missing parts of the task dynamics.
At the same time, \textbf{TransPrec decreases only mildly}, indicating that the learned task graph remains structurally meaningful even as the dataset grows. At $|\mathcal{D}|=30$, no feasible abstraction was found with the tested $(k_{\text{pick}},k_{\text{place}})$ value, indicating that the trace is too sparse to support a graph that simultaneously satisfies bipartiteness and action-uniqueness.
Finally, we report runtime as a function of $|\mathcal{D}|$ (\autoref{fig:scalability}, bottom).
Computing the visual-distance heuristic dominates the cost at larger $|\mathcal{D}|$, while constrained search time increases with the size of the induced graph.

Finally, we validate VGGC on two real-robot benchmarks (\autoref{tab:real_robot}).
Despite sensor noise and real-world variability, VGGC achieves \textbf{94\%} and \textbf{70\%} image-to-plan success.
\section{Conclusions}


We presented \emph{Vision-Guided Graph Coloring (VGGC)}, a framework that constructs discrete, action-consistent task graphs directly from RGB execution traces with discrete action labels.
By enforcing structural constraints during graph construction and using vision only to break ties among feasible solutions, VGGC yields abstractions that are immediately usable for high-level planning.

Across simulation and real-robot experiments, VGGC achieves high plan success and scales to larger region/object settings than clustering-only baselines.
A limitation is that VGGC relies on observed transitions; incorporating learned forward dynamics or structured factorization could improve generalization to unseen states.
Future work will explore more scalable constrained solvers and factored/hierarchical abstractions that preserve these structural guarantees.

\bibliographystyle{IEEEtran}
\bibliography{refs, references}

\end{document}